\documentclass{article}

\usepackage{microtype}
\usepackage{graphicx}
\usepackage{subfigure}
\usepackage{booktabs} % for professional tables

\usepackage{hyperref}

\usepackage[accepted]{icml2020}

\usepackage{amsmath}

\icmltitlerunning{Submission and Formatting Instructions for ICML 2020}

\newcommand{\bl}[1]{\textcolor{black}{#1}}

\begin{document}

\twocolumn[
\icmltitle{Enabling Deep Learning on Edge Devices through Filter Pruning and Knowledge Transfer}




\begin{icmlauthorlist}
\icmlauthor{Kaiqi Zhao,}{}
\icmlauthor{Yitao Chen,}{}
\icmlauthor{Ming Zhao}{}\\
\icmlauthor{Arizona State University}{}
\end{icmlauthorlist}

% \icmlaffiliation{}{Department of Computation, University of Torontoland, Torontoland, Canada}
% \icmlaffiliation{go}{Googol ShallowMind, New London, Michigan, USA}
% \icmlaffiliation{end}{School of Computation, University of Edenborrow, Edenborrow, United Kingdom}

% \icmlcorrespondingauthor{Cieua Vvvvv}{c.vvvvv@googol.com}
% \icmlcorrespondingauthor{Eee Pppp}{ep@eden.co.uk}

\icmlkeywords{Machine Learning, ICML}

\vskip 0.3in
]



\pagestyle{plain}
% !TEX root = main.tex
\begin{abstract}
Deep learning models have introduced various intelligent applications to edge devices, such as image classification, speech recognition, and augmented reality. There is an increasing need of training such models on the devices in order to deliver personalized, responsive, and private learning. To address this need, this paper presents a new solution for deploying and training state-of-the-art models on the resource-constrained devices. First, the paper proposes a novel filter-pruning-based model compression method to create lightweight trainable models from large models trained in the cloud, without much loss of accuracy. Second, it proposes a novel knowledge transfer method to enable the on-device model to update incrementally in real time or near real time using incremental learning on new data and enable the on-device model to learn the unseen categories with the help of the in-cloud model in an unsupervised fashion. The results show that 1) our model compression method can remove up to 99.36\% parameters of WRN-28-10, while preserving a Top-1 accuracy of over 90\% on CIFAR-10; 2) our knowledge transfer method enables the compressed models to achieve more than 90\% accuracy on CIFAR-10 and retain good accuracy on old categories; 3) it allows the compressed models to converge within real time (three to six minutes) on the edge for incremental learning tasks; 4) it enables the model to classify unseen categories of data (78.92\% Top-1 accuracy) that it is never trained with. 
\end{abstract}
% !TEX root = main.tex
\section{Introduction}\label{introduction}

Deep neural networks (DNNs) have been applied to many important applications on edge devices, such as image classification, speech recognition, and augmented reality. These deep learning models typically have millions of parameters and need to be trained for hours or even days on powerful cloud servers to achieve a good performance. However, a serious drawback of this cloud-only approach is that the on-device tasks cannot perform well when the cloud is overloaded or the network is unreliable. Moreover, there are also significant benefits from training deep leaning models on edge devices: 1) Customization: user- or situation-specific requirements can be met more effectively by training models on the devices that the users or physical environments directly interact with; 2) Responsiveness: custom models deployed on devices for specific users or environments can better adapt to their changing behaviors using new data captured by the devices; and 3) Privacy: sensitive information can be better protected if the sensitive data and models are stored and used only on private devices, not in the public resources shared by many.
	
Deploying and training complex deep learning models on edge devices are challenging since \bl{they require millions of parameters and large amounts of operations whereas the devices have only limited memory and computation resources}. To deploy DNNs on resource-constrained devices, there are two general approaches. The first approach aims to compress already-trained models, using techniques such as weights sharing~\cite{chen2015compressing}, quantization~\cite{han2015deep, kadetotad2016efficient}, and pruning~\cite{han2015deep,lecun1990optimal,srinivas2015data}. However, a compressed model generated by these approaches is useful only for inference; it cannot be re-trained to capture user- or device-specific requirements or new data available at runtime.

The second approach to learning on devices is based on knowledge transfer  which uses the knowledge distilled from a cloud-based deep model (termed teacher) to improve the accuracy of a on-device small model (termed student)~\cite{ba2014deep,hinton2015distilling,romero2014fitnets,venkatesan2016diving}. However, these works 1) achieve limited accuracy improvement~\cite{yim2017gift,zagoruyko2016paying}; 2) do not consider the speed of training the model to a satisfactory accuracy; and 3) assume that the all data are available at the training time and the tasks for the student and teacher remain exactly the same, which is often not a realistic assumption. %Further, the student may have access to local data that exceeds the scope of the teacher's knowledge. Once the student model is trained with the new data, its performance on the old tasks degrades significantly, which is termed catastrophic forgetting. Inspired by prior work~\cite{li2017learning}, we use knowledge distillation and group loss to reduce the forgetting.

The goal of our work is to provide a new solution that allows deep learning models to be trained on devices with a small number of parameters, the state-of-the-art accuracy, and fast runtime. Further, we aim to enable on-device learning under realistic settings where the models are trained incrementally with only limited local input but are still able to recognize both old and new categories of data.

In order to achieve the above goal, we propose a new compression method for deploying models that are suitable and trainable for resource-constrained devices, and a new knowledge transfer method for improving the training accuracy and the speed of these on-device models, and providing the capability for enabling the on-device models to learn incrementally without forgetting the knowledge on the old categories using the local data and achieve good accuracy for classifying both old and new categories. Specifically, our compression method can create a model that is both shallow and thin by removing similar convolution layers and pruning filters that produce weak activation patterns in each layer, respectively, from a large model trained in the cloud. The resulting compressed model still shares the same architecture as the original model, and is suited for knowledge transfer between the two models. \bl{Our proposed knowledge transfer method selects the best teacher/student layer pairs for transferring knowledge from teacher's intermediate representations and enables the student to learn the problem solving process.} Our proposed method also enables the student to use the distilled knowledge from the teacher in solving the catastrophic forgetting problem. 
% As a result, the student is able to learn from the new local data without forgetting its old knowledge, and classify data from categories that it has never seen before. 

%Our proposed knowledge transfer method identifies the best teacher-student pairs for knowledge transfer using a cosine similarity metric, and transfers knowledge by minimizing the loss functions built from the outputs of the layer pairs without using the input's true labels

We evaluate our solution on VGG-16 and ResNet architectures using CIFAR-10, Caltech 101, and ImageNet datasets. First, our model compression method 1) reduces 99.36\% parameters of WRN-28-10, while preserving a Top-1 accuracy of over 90\% on CIFAR-10; and 2) achieves a compression ratio of up to 139X on VGG-16, at a cost of less than 10\% accuracy loss on Caltech 101. \bl{Second, our knowledge transfer method 1) enables the compressed models not only perform well on new category ($>$90\% accuracy on CIFAR-10) but also retains a good level of accuracy for classifying the old categories; and 2) enables the compressed models to converge within real time (three to six minutes) on the edge for incremental learning tasks; and 3) allows the compressed model to reach a Top-1 accuracy of 78.92\% on CIFAR-10 for classifying unseen categories that it is never trained with. Compared to the related works~\cite{romero2014fitnets,zagoruyko2016paying}, our method reduces complex networks to both shallower and thinner networks without much loss of accuracy, enables the models to learn from new categories incrementally within real time without forgetting the old categories, and allows the models to classify unseen categories of data with both good accuracy and speed.}

In summary, our solution enables DNNs that are not only suitable for deployment on resource-constrained devices but also trainable for meeting new requirements. In the rest of the paper, we first explain the details of our proposed solution (Section 2), then present an extensive evaluation (Section 3), discuss the related works (Section 4), and finally conclude the paper (Section 5).
% !TEX root = main.tex
\section{Background and Motivations}
We envision an edge computing scenario where edge devices collect various data (voice, images, videos, etc.) from their sensors and feed it to the cloud. In the cloud, we can utilize the abundant resources in the cloud to train a state-of-the-art model with all the available data. On the edge, we can deploy a  small model on each device and train it using the local data for customized, responsive, and private learning. 
% What are the requirements for this architecture to work? 
% 1. Edge model needs to be small enough
% 2. On-device model should be able to categorize unseen categories. 
% 3. On-device model should be able to categorize new categories without forgetting old categories. 

\bl{
In order to realize the above scenario, the cloud/edge distributed learning system needs to meet the following requirements. First, the on-device model should be small enough to fit the limited resources on the edge devices, which are usually resources constrained due to their small form factor. Second, the on-device models should be able to classify new categories without forgetting old categories since re-training the whole model on edge devices is infeasible due to their limited computing resources. Third, the on-device model should be able to classify unseen categories with good accuracy, since each edge device may only see a subset of the data that the cloud model is trained with. } 
% In order to meet these requirements, we propose a novel model compression and knowledge transfer techniques, which takes the following steps iteratively. 
% need for model compression

% It is crucial to compress the cloud model before the deployment due to the limited resources on edge devices. Compressed models need less storage space and more importantly, they can run faster than the uncompressed models due to the reduction in the number of parameters. Knowledge transfer techniques allow our on-device models to learn from the in-cloud model without forgetting the old data that the in-cloud model is already trained with and to achieve real time or near real time update since the on-device models can now learn incrementally on new data. In addition, knowledge transfer techniques allow the on-device models to learn the unseen categories with the help of the in-cloud model, but without relying on data labels which may not be available to this device. 
To meet the above requirements, we need to use compression techniques to produce models that are small enough and fast enough for the edge devices with their limited resources. We also need knowledge transfer techniques that can utilize the knowledge of the in-cloud model to help the on-device models retain the existing knowledge while learning on new data and be able to classify categories that they are not trained with.

But on one hand, existing model compression methods focused only on creating compressed models for efficient inference without considering how to compression methods affect the training process~\cite{han2015deep, chen2015compressing, kadetotad2016efficient, li2016pruning, polino2018model}, and how to reduce the accuracy loss caused by compression. On the other hand, existing knowledge transfer methods have the following limitations: 1) they still require large student models that are not fit for resources constrained devices~\cite{romero2014fitnets,li2019rilod, yim2017gift}; 2) they only enable to student model to classify the categories that the models are trained with.   

% Hence, we propose a novel model compression method to create an on-device model that fits the device's limited resources with low accuracy loss. In addition, our compressed models are trainable, which is key to training customized, responsive, and private models on edge devices. our selective layer-wise knowledge transfer method enables the on-device model to reuse the already exist knowledge from the cloud model to classify the unseen categories without using label, i.e., in an unsupervised fashion, for a faster convergence speed than learning by itself.  

To address these limitations and meet the aforementioned requirements, we propose novel model compression and knowledge transfer techniques for deploying models that are suitable and trainable for resource-constrained devices and improving the training accuracy and speed of these on-device models, as detailed in the rest of the paper.
\section{Filter Pruning Based Model Compression}
\begin{figure}[t]
	\centering
		\subfigure[Techer-Student ResNet]{
		\begin{minipage}[t]{0.5\linewidth}
			\centering
			\includegraphics[width=1.7in]{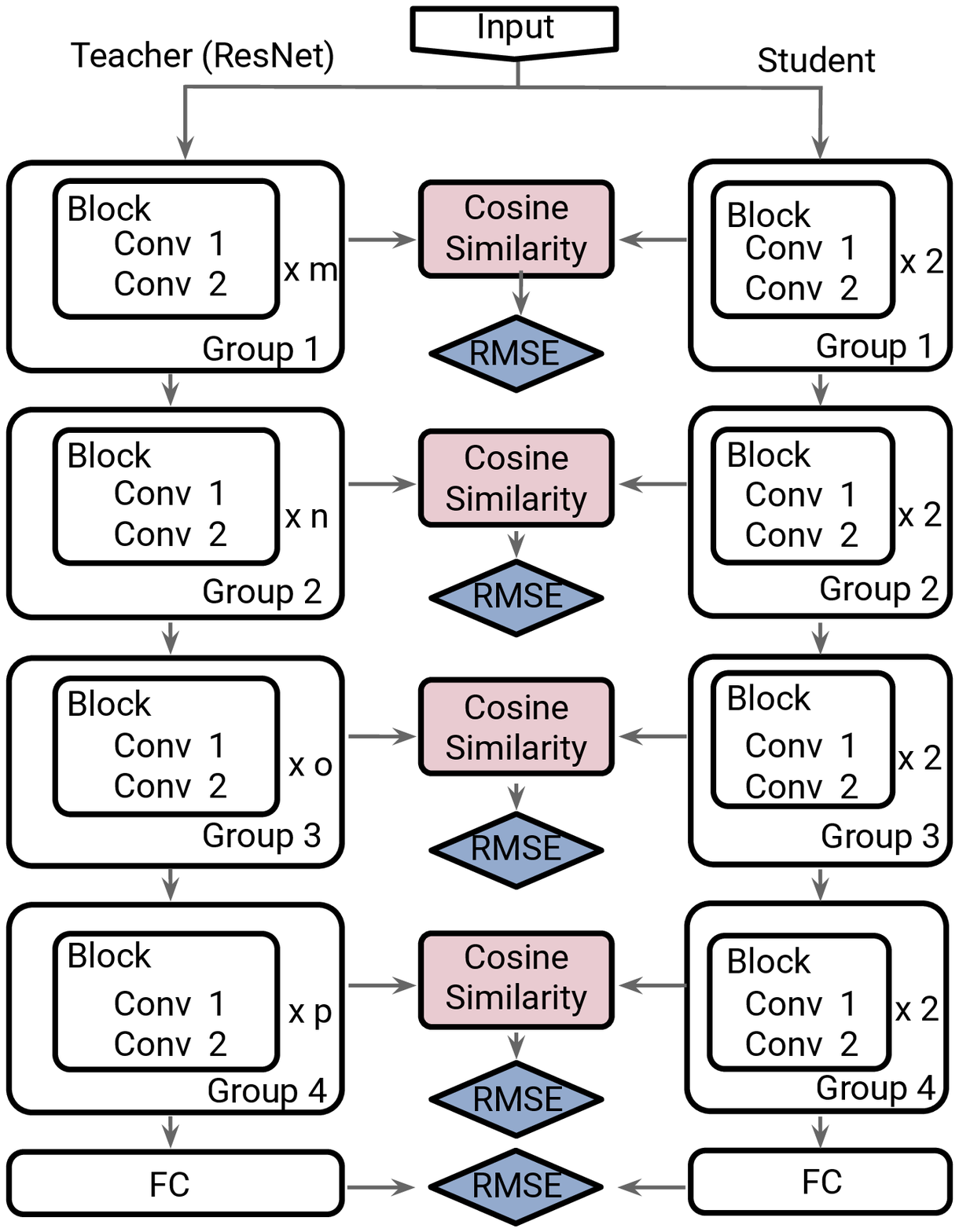}
		\end{minipage}%
	}%
	\subfigure[Techer-Student VGG-16]{
		\begin{minipage}[t]{0.5\linewidth}
			\centering
			\includegraphics[width=1.45in]{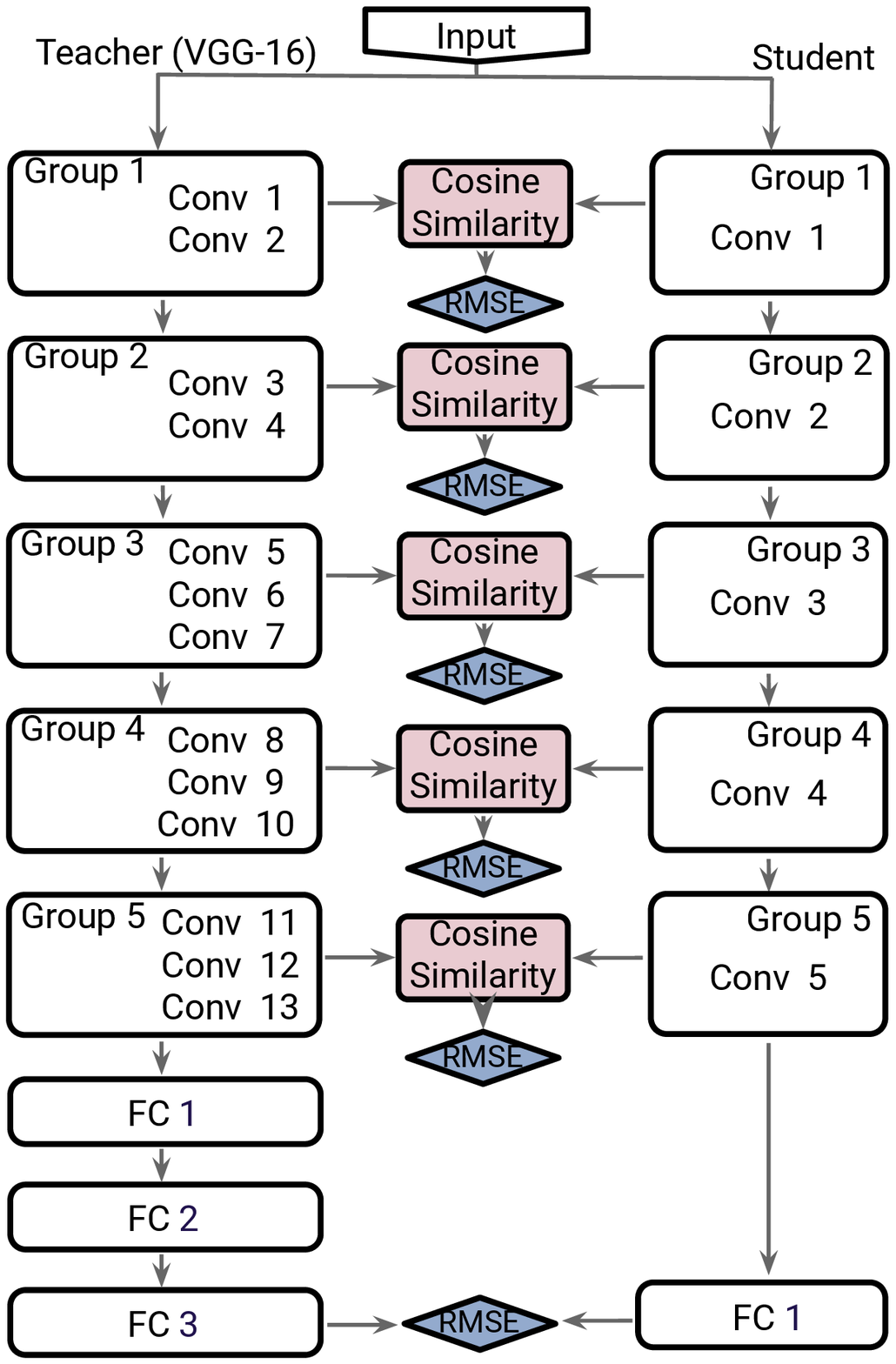}
		\end{minipage}%
	}%
   \centering
   \caption{Schematics of the proposed knowledge transfer method for ResNet models (on the left) and VGG-16 models (on the right). 
%   m, n, o, p denote the number of blocks in each group of the ResNet model. For ResNet-18, they are 2,2,2,2; for ResNet-34, they are 3,4,6,3.
   }
   \vspace{-0.2in}
   \label{Schematics}
\end{figure}

Without loss of generality, we consider image classification tasks and use ResNet, as an example to discuss our proposed on-device learning solution. Image classification is important for many edge applications, and is also the target task of the related model compression and knowledge distillation works~\cite{hinton2015distilling, han2015deep, chen2015compressing, polino2018model, srinivas2015data}. ResNet is a modern architecture with streamlined convolutional layers. Specifically, we consider WRN-28-10 and ResNet-34, illustrated as Teacher in Figure~\ref{Schematics}. They have a Top-1 accuracy of 97.28\% on CIFAR-10 and 73.9\% on ImageNet, respectively, which are among the state-of-the-art results. The ResNet models consist of several groups of blocks, and each block has two convolutional layers. Further, we also consider VGG-16~\cite{simonyan2014very}, which is another commonly used neural network and has a different architecture, including 13 convolutional layers and three fully-connected layers.

Our goal for model compression is two-fold: 1) to reduce the number of parameters and optimize the architecture of the model so that it is both thin and shallow, and fit for the limited resources on a device; 2) to maintain the architecture of the original model so that it can facilitate the learning from the on-server model during knowledge transfer.

The proposed model compression method works as follows. First, to reduce depth, it creates a shallower model that has the same number of groups as the on-server model, but each group only keeps the last block (illustrated as Student in Figure~\ref{Schematics}). This way of pruning layers of a model also resonates with the principle that higher layer features are closer to the useful features for performing a main task~\cite{yim2017gift}. Next, our method reduces the width of the shallower model by removing filters that produce weak activation patterns. It uses one batch of images to decide the number of filters that are safe to prune in each convolution layer.
%
%For each convolution layer, we count the total number of filters that are safe to prune for each input image, and calculate its average value for one batch of images; then we use the average value to determine the number of filters that can be pruned, and finally retrain the model with remaining filters. 

The procedure of pruning filters from the $i$th convolution layer is as follows. \bl{For a given input image m, let $X^m_{i-1}$ denote the input features of the $i$th convolution layer.} Convolution operations (denoted as mapping function $F$) transform the input $X_{i-1}^m$ into output feature maps $Z_{i}^m$ by applying $n_{i}$ three-dimensional filters $f_{i,j}^m$. Then, activation operations (denoted as mapping function $G$) transform $Z_i^m$ into the activation feature maps $A_i^m$:
\begin{equation}
Z_i^m = F (X_{i-1}^m),\quad A_i^m = G (Z_i^m)
\end{equation}
For each filter's activation feature map $a^m_{i,j} \in R^{h_i \times w_i} $ ($1 \leq j \leq  n^{i}$), our method computes the percentage of zero elements based on the $l_0$-norm of $a^m_{i,j}$:
\begin{equation}
perc(a^m_{i,j}) = 1 - \frac{{\| a^m_{i,j} \|}_0}{h^{i} \times w^{i}}.  
\end{equation}
If the percentage is equal to or greater than Filter Pruning Threshold $P$, this filter is safe to be pruned. The threshold determines how aggressive the pruning is, and in the evaluation, we set it \bl{between 0.7 and 1.0}. \bl{Our method repeats the above procedure for M randomly selected images, and calculates the average number ($avgc_i$) of filters that are safe to prune.} We set $M$ equal to batch size since we find that the value of $avgc_i$ is steady even if the input features are different. The reduced width $w^i$ of the $i$th convolution layer \bl{becomes}: $w^i=n^i-avgc_i$. \bl{The same method is applied to all the remaining layers of the shallow model.} The model is then retrained with the reduced width and depth to generate the compressed on-device model.

%As $avgc_i$ filters of the $i$th layer are pruned, the kernels of the next convolution layer corresponding to the pruned feature maps are also removed, and a new kernel matrix is generated for both the $i$th and ($i+1$)th layers

\begin{figure}[t]
	\centering
	\includegraphics[width=0.45\textwidth]{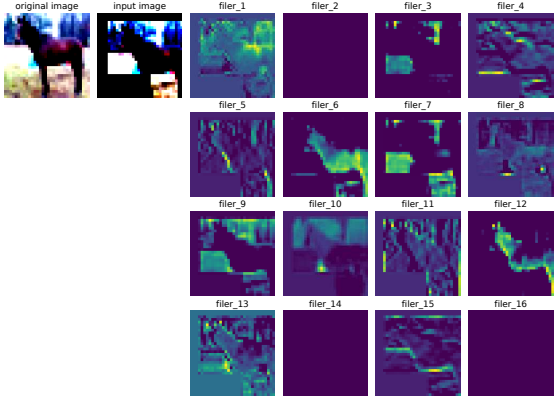}
	\caption{Activation feature maps of each filter of the first convolutional layer (Conv1) of the on-server model (WRN-28-10) on CIFAR-10. In the left, the first image shows the original image and the second image shows the input features after data augmentation; the right part shows activation features of the 16 filters.}
	\label{visualization_of_activations}
	\vspace{-0.15in}
\end{figure}

We can visualize the activations of the on-server model (WRN-28-10) on CIFAR-10 and understand why our filter pruning method is effective. Figure~\ref{visualization_of_activations} shows the activation feature maps of each filter of the first convolutional layer (called Conv1) using one image as the input. The width of Conv1 is $16$. The first image on the left is the original image, and the second image is the input features after data augmentation. We can see that some filters extract lots of representations with high activation patterns, like the $6$th and $12$th filters, whereas the activation feature maps of some filters are close to zero, such as the $2$nd, $14$th, and $16$th filters. Filters that generate weak activations are safe to remove without affecting the final performance of the model.

In this way, we can generate a compressed model that is both shallow and thin, small enough for learning on edge devices. The small model still shares the same architecture of the original model, because it retains the higher layers in each group of convolutional layers and keeps important filters in each remaining layers. Compared to the related filter pruning work that prunes filters with the lowest absolute weight sum~\cite{li2016pruning}, our approach prunes insignificant filters more accurately. Filters that have small absolute weight sum can also produce useful non-zero activation patterns that are important for learning features during backpropagation. \bl{As shown in Table~\ref{tab:model_compression} of Section~\ref{model_compression_results}, our method enables the compressed model achieving a higher Top-1 test accuracy than their method (93.68\% vs 93.55\%), with a smaller number of parameters (1.42M vs 1.68M). So} our approach directly finds and prunes the filters that generate close-to-zero activations, with minimal impact on the performance. 

% !TEX root = main.tex
\section{Selective Layer-Wise Knowledge Transfer}
\subsection{Knowledge Transfer for Incremental Learning on the Edge}
\bl{As new local input becomes available to a device, we want to update its local model to learn the new data. One solution is to wait for the in-cloud model to update using all the data from the edge and then compress and download the updated model, which however will take a significant amount of time. In order to update the on-device model in real time or near real time, we propose to update it incrementally using its new data, and use knowledge transfer from the in-cloud model (the teacher) to preventing the on-device model (the student) from forgetting the old data that it is already trained with.}

Different from the existing works~\cite{kim2018paraphrasing,romero2014fitnets, zagoruyko2016paying}, with our knowledge transfer method, the student does not need to learn the specific output from the teacher, which depends on the specific input; it instead learns the problem solving process, which represents the intermediate layer outputs. Learning from the teacher's intermediate representations is better than learning from only the last layer's output~\cite{sharma2018existing}, which prevents the model from losing its classification ability when facing specific questions.

Figure~\ref{Schematics} illustrates the architecture of our knowledge transfer method for ResNet and VGG-16. The student is trained by knowledge transfer between selected teacher-student layer pairs as the input enters batch by batch at each iteration. First, given one batch of data, our method finds out which convolutional blocks in the teacher should be used to transfer knowledge to the student's convolutional blocks, using a new cosine similarity based metric. Then, multiple loss functions are built using the activations from the mapped block pairs.

In order to find best teacher-student layer pairs for knowledge transfer, first, we define a cosine similarity metric for measuring the similarity between the activation feature maps of the teacher's $k$th block and the student's $j$th block: 
\begin{align}\label{cosine}
    &CosineSim_{k,j}(X)=\frac{ Q^t_{k} \cdot Q^s_{j}}{ \left\| Q^t_{k}\right\| \left\| Q^s_{j}\right\|}, \\
    &Q^t_{k} = \frac{F^t_k(X)}{ \left\| F^t_k(X)\right\|},  \quad Q^s_{j} = \frac{F^s_j(X)}{  \left\| F^s_j(X)\right\|},
\end{align}

where,\\
$X$: one batch of data.\\
$F^t_k(X)$: activation feature maps of the teacher's $k$th block.\\
$F^s_j(X)$: activation feature maps of the student's $j$th block.\\

As shown above, the cosine similarity is calculated using $l_2$-normalized feature maps, which helps the student's learning by normalizing activations of the teacher and student into a similar scope. In addition, our method does zero padding on the activation features maps of the student models before normalization, since the width of convolution layers of compressed student models is different from that of the teacher. It calculates the cosine similarity between each pair of teacher/student blocks in the same group, and the pairs $(k^*,j^*)$ that produce the largest cosine similarity value are mapped together for knowledge transfer.

%First, mapped blocks are selected in the same way as the model compression and edge deployment stage discussed in the previous section. Then the loss function is built by adding all the loss terms from intermediate layers and the fully-connected layers together, using $J_1$ and $J_b$ in Eq.~\ref{loss_function1} as follows:

Then the loss function is built by adding all the loss terms from intermediate layers ($J_b$), the fully-connected layers ($J_1$), and the cross entropy loss ($J_3$) with true labels of the dataset together, defined as follows:

% \begin{align*}\label{loss_function1}
% 	\ell &= \lambda_1J_1 + \lambda_2\sum_{m=1}^gJ_b 
% 	+ \lambda_3J_3\\
% 	J_1(FC^t, FC^s) = \sqrt{ \sum_{i=1}^{n} (FC^t_{i}-FC^s_{i})^2 };\\
%     J_b(Q^t_k, Q^s_j) = \sqrt{\sum_{i=1}^{n} (Q^t_{ki}-Q^s_{ji})^2}, \quad b=1,2,3,...,g;\\
% 	J_3 &= \sum_{i=1}^{c} [Y_{i}log\widehat{Y_{i}^s} + (1 - Y_i)log(1 - \widehat{Y_{i}^s})]\\
% \end{align*}

\begin{align*}\label{loss_function1}
	&\ell = \lambda_1J_1 + \lambda_2\sum_{m=1}^gJ_b + \lambda_3J_3\\
	&J_1(FC^t, FC^s) = \sqrt{ \sum_{i=1}^{n} (FC^t_{i}-FC^s_{i})^2 };\\
    &J_b(Q^t_k, Q^s_j) = \sqrt{\sum_{i=1}^{n} (Q^t_{ki}-Q^s_{ji})^2}, \quad b=1,2,3,...,g;\\
	&J_3 = \sum_{i=1}^{c} [Y_{i}log\widehat{Y_{i}^s} + (1 - Y_i)log(1 - \widehat{Y_{i}^s})]\\
\end{align*}

where,\\
$\qquad \lambda_1, \lambda_2, \lambda_3$: hyper-parameters to balance the weights of different loss terms\\
$\qquad c$: the number of classes of the datasets\\
$\qquad g$: the number of groups of the teacher/student\\
$\qquad n$: the number of feature maps of the teacher/student\\
$\qquad FC^t$: output of teacher's last fully-connected layer\\
$\qquad FC^s$: output of student's last fully-connected layer\\
$\qquad Q^t_k$: $l_2$-normalized output of teacher's $k$th block\\
$\qquad Q^s_j$: $l_2$-normalized output of student's $j$th block\\
$\qquad \widehat{Y^s}$: predicted softmax output of the student\\
$\qquad Y$: true labels of the datasets\\

Note that, as the input changes batch by batch, the mapped block pairs also change according to the cosine similarity, in order to ensure knowledge transfer is always done with the best teacher-student pairs. During backpropagation, our method only updates the weights of the last convolutional layer in each group of the student model while minimizing the loss function. This form of updating is reasonable since: 1) freezing some of the layers corresponding to the original model can help limit its adaptability to new data\cite{jung2016less,castro2018end}; 2) higher layer features are closer to the useful features for performing a main task~\cite{yim2017gift}; and 3) updating less layers allows the training to complete sooner on resource-constrained devices.

\subsection{Knowledge Transfer for Classifying Unseen Categories on the Edge}

\bl{As the in-cloud model improves over time from the data fed by the edge, there are unseen categories for the on-device models since each edge has seen only a subset of that data that the cloud is trained with. Given the data belonging to the unseen categories, the on-device model cannot classify it but the in-cloud model can. One way to solve this problem is to compress the in-cloud model and download it again to the device. Alternatively, we propose to also use the aforementioned knowledge transfer method to enable the existing model (the student) on the device to learn the unseen categories, with the help from the in-cloud model (the teacher), but without relying on data labels which may not be available to this device.}

\bl{First, mapped blocks are selected in the same way as the knowledge transfer for classifying unseen categories discussed in the previous section.} Next, the loss function is built by using only the mapped block pairs of all the groups ($J_b$) and the last fully-connected layer of the teacher and student models ($J_1$) from Eq.~\ref{loss_function1}.

% !TEX root = main.tex
\section{Evaluation}\label{evaluation}
%We first evaluate the effectiveness of our filter pruning based model compression method for creating lightweight compressed models. Then we evaluate our knowledge transfer method for further improving the performance of compressed models.
%\bl{The evaluation of our cloud-edge system focuses on answering the following questions: 1) How effective is the proposed model compression method for creating lightweight compressed models? 2) How effective and efficient is the selective layer-wise knowledge transfer method for classifying unseen categories and incremental learning tasks on the edge? 3) How does this complete system work in practice? What is the runtime?}

%\bl{We consider four important performance metrics: 1) Compresssion ratio is the proportion of the number of parameters of the compressed on-device model and that of the on-server model; 2) Classification accuracy is the proportion of correctly predicted labels among all the predictions obtained by the network. In Top-1 accuracy, the predicted label is counted as correct when the label with the highest probability equals the target label; 3) Convergence speed is measured as the total number iterations required by the model to reach 90\% of the Top-1 test accuracy; 4) Runtime is evaluated by the total time required of the model to converge on the edge device.}

We implemented our solution on TensorFlow version r1.3, and elvaluated the cloud model on a Nvidia Tesla K40 GPU, hosted on a server equipped with dual Intel Xeon E5-2630 processors and 64GB of main memory. We evaluate our edge model on a commercialized device, Google Pixel 2, which has an eight-core, Qualcomm Kryo 280 CPU and 4GB of main memory. 
In our experiments of ResNet models, we used SGD with Nesterov momentum for optimization. Dampening was set to 0, momentum to 0.9, initial learning rate to 0.1, and mini-batch size to 128. 
On CIFAR-10, weight decay was set to to 0.0005, and learning rate decayed each epoch with the cosine annealing schedule, training for total 200 epochs; 
On ImageNet, weight decay was set to to 0.0001, and learning rate dropped by 0.1 at 30, 60, and 90 epochs, training for total 100 epochs. 
In the experiments of VGG-16 models, we used Adam for optimization. Initial learning rate was set to 0.01 for the original on-server model and 0.001 for compressed models. They decayed exponentially each epoch with a factor of 0.98. Validation and test accuracy of all the models were calculated at each epoch. Final accuracy of the model was reported as the test accuracy attained at the epoch with the highest validation accuracy.
%In our experiments of ResNet models on CIFAR-10, we used SGD with Nesterov momentum for optimization. Weight decay was set to to 0.0005, dampening to 0, momentum to 0.9, and mini-batch size to 128. The initial learning rate was 0.1 and decayed each epoch with the cosine annealing schedule.

%\bl{We compare our novel model compression and knowledge transfer techniques with state-of-the-art related works. For model compression, we compare with three methods: a related filter pruning method that prunes filters with the smallest sum of absolute kernel weights~\cite{li2016pruning}, PM (“post-mortem”) Quantization~\cite{polino2018model}, and Quantized Distilition~\cite{polino2018model}. We also compare the proposed knowledge transfer method with FitNet~\cite{romero2014fitnets}, Attention Transfer~\cite{zagoruyko2016paying}, and Factor Transfer~\cite{kim2018paraphrasing}. Further, to evaluate the performance of our model on learning incremental data, we compare with two recent state-of-the-art methods, Global Distillation (GD) and RILOD.}

%\subsection{Benchmark Datasets and Data Augmentation}
We conducted experiments on three important datasets:

\textbf{CIFAR-10} consists of 60,000 (32X32) RGB natural images, belonging to 10 classes with 6000 images per class~\cite{krizhevsky2009learning}. Each image is 32X32 pixels in 3 color channels.
% There are 50,000 training images and 10,000 test images. 
% , yielding input vectors with 3072 dimensions.

\textbf{Caltech 101} consists of 9145 (224X224) RGB images from 101 classes. Each class has 40 to 800 images. We divide the dataset into three parts: the training set consists of 5853 images (64\% of the total dataset), the testing set consists of 1829 images (20\%), and the validation set consists of 1463 images (16\%)~\cite{fei2007learning}.

\textbf{ImageNet} consists of over 14 million RGB images organized into 21,841 classes. Each class has over 500 images. We use the subset of images with SIFT features, which belong to 1000 classes~\cite{deng2009imagenet}.
% The number of images with SIFT features is 1.2 million, belonging to 1000 classes~\cite{deng2009imagenet}.

We preprocess all data by subtracting the mean and dividing by the standard deviation of each image vector. For experiments on CIFAR-10 and Caltech 101, all training images are padded 4 pixels on each side, and a 32X32 crop is randomly sampled from the padded image. Then the images are flipped left-right randomly with a probability of 0.5 and masked out randomly with a cutout size of 16X16 pixels~\cite{lee2015deeply}. For experiments on ImageNet, all training images are first cropped randomly with a size of 224X224, and then horizontally flipped randomly with a given probability of 0.5. 
% For testing images, we only evaluate the single view of the original image. This augmentation exposes the model to additional variations without the cost of collecting more data, and thereby improves the model's ability to generalize~\cite{romero2014fitnets}.

% In the rest of this section, we evaluate 1) model compression in terms of the size and accuracy of the compressed model for edge deployment, and 2) knowledge transfer in terms of the accuracy and speed of the on-device model. 

\subsection{Results for Model Compression}\label{model_compression_results}
\begin{table}[t]
	\centering
	\caption{Model compression results of WRN-28-10, VGG-16, and ResNet-34 on CIFAR-10, Caltech 101, and ImageNet, respectively. P denotes pruning threshold.}
	\label{tab:model_compression}
	\small
	\begin{tabular}{p{1.2cm}p{1.57cm}p{0.3cm}p{1.0cm}p{1.15cm}p{0.7cm}}
		\toprule
		Data & Model & P & Accuracy & Param-  & {Comp.} \\
		Set  & Name  &   &      & eters (M)    & Ratio\\
		\hline
		         & WRN-28-10 &  & \textbf{97.28\%} & 36.22 & \\
		         & ResNet 1 & 1.0 & 94.37\% & 2.00 & 18$\times$\\
		CIFA- & ResNet 2 & 0.9 & 93.68\% & 1.42 & 25$\times$\\
		R10         & ResNet 3 & 0.8 & 92.62\% & 0.60 & 60$\times$\\
		         & ResNet 4 & 0.7 & 90.09\% & \textbf{0.23} &\textbf{160$\times$}\\
	    \hline
		         &ResNet-34   &   & \textbf{73.23\%} & 21.6 & \\
		         &ResNet 5 & 1.0 & 69.76\% & 9.79 & $2\times$\\
		Image-   &ResNet 6 & 0.9 & 68.14\% & 7.22 & $3\times$\\
		Net      &ResNet 7 & 0.8 & 66.07\% & 5.09 & $4\times$\\
		         &ResNet 8 & 0.7 & 63.16\% & \textbf{3.34} & \textbf{$6\times$}\\
		\hline
		            &VGG-16   &   & \textbf{77.10\%} & 134 & \\
		            &VGG-16 1 & 1.0 & 62.85\% & 5.55 & 24$\times$\\
		Calt- &VGG-16 2 & 0.9 & 60.55\% & 3.78 & 36$\times$\\
		ech 101             &VGG-16 3 & 0.8 & 59.51\% & 3.11 & 43$\times$\\
		            &VGG-16 4 & 0.7 & 56.77\% & \textbf{0.97} &\textbf{139$\times$}\\
		\bottomrule
	\end{tabular}
	\vspace{-0.2in}
\end{table}

%The network parameters and accuracy before and after pruning different numbers of filters are shown in Table~\ref{tab:model_compression}. 
We first experiment on CIFAR-10 dataset with ResNet (WRN-28-10) as the on-server model. By changing the pruning threshold P, our method can flexibly generate four compressed models, ResNet 1-4, offering different tradeoffs between size and accuracy, shown in Table~\ref{tab:model_compression}. The results show that all the compressed models can achieve good compression ratios without losing much accuracy. \bl{In particular, compressed ResNet 4, the size of which is only 0.64\% of the origin model WRN-28-10, still remains a Top-1 accuracy of over 90\%. ResNet 7 achieves a compression ratio of 4X at the cost of 7.16\% loss in accuracy. The compressed model VGG-16 4 achieves a compression ratio of up to 139X at the cost of less than only 10\% loss in accuracy.}

Table~\ref{tab:Comparison of model compression methods} shows the comparison of the proposed model compression method and the related works on CIFAR-10. The related filter pruning work~\cite{li2016pruning} ResNet-110-prune was evaluated on ResNet-110, and the related PM (“post-mortem”) quantilization and quantized distillation works were evaluated on WRN-28-20. Our method allows the compressed ResNet 2 (93.68\%) to achieve a comparable accuracy as that of ResNet-110-pruned~\cite{li2016pruning} (93.55\%), quantized distillation (94.73\%), and a higher accuracy than PM quantization~\cite{polino2018model} (81.09\%) while requiring much less parameters (1.42M) than all these four compressed models (5.4M, 1.68M, 7.44M, and 9.66M). Meanwhile, the compression ratio of compressed ResNet 2 (25X) also outperforms that of all other compressed models (3X, 1X, 19X, and 15X) significantly, achieving a much higher compression ratio and producing a much smaller model for edge deployment.

%\bl{Compared to quantized distillation, compressed ResNet 1 and compressed ResNet 2 need 1.42M and 2M parameters, respectively, which are about 85\% and 80\% less than the number of parameters necessary for quantized distillition (9.66M). At the same time, the accuracy of these three compressed models are all about 94\%. Note that our compressed models are created from a smaller model (WRN-28-10), which is more difficult than the one (WRN-28-20) that quantized distillation are based on---WRN-28-10 (36.22M) is only 25\% of that of WRN-28-20 (145M). But even in this context, our method can provide more compression than quantized distillation.} 

\begin{table}[t]
	\centering
	\caption{Comparison of the proposed model compression method on CIFAR-10.}
	\label{tab:Comparison of model compression methods}
	\begin{tabular}{p{2.7cm}p{1.2cm}p{1.6cm}p{1.0cm}}
		\toprule
		Model Name & Accuracy & Parameters (M) & Comp. Ratio\\
		\hline
		WRN-28-10 & \textbf{97.28\%} & 36.22 & \\
		ResNet 1  & 94.37\% & 2.00  & 18$\times$\\
		ResNet 2  & 93.68\% & \textbf{1.42}  & \textbf{25$\times$}\\
% 		VGG-16         & 93.25\% &  15  & \\
% 		VGG-16-pruned& 93.4\%  & 5.4  & 3$\times$\\
		\hline
		ResNet-110& 93.53\%  &1.72  & \\
		ResNet-110-prune& 93.55\%  &1.68  & 1$\times$\\
		\hline
		WRN-28-20   & 95.74\%  & 145 & \\
		PM Quantization   & 81.09\%  & 7.44  & 19$\times$\\
		Quantized Distill.& 94.73\%  & 9.66  & 15$\times$\\
		\bottomrule
	\end{tabular}
	\vspace{-0.1in}
\end{table}

\subsection {Knowledge Transfer}
We use on-server models (WRN-28-10 and VGG-16) as the teacher model, and their corresponding compressed models as the student model. We compare the performance of the student model that is trained with the help from the teacher, called the dependent student, with two baselines: the teacher
model and the independent student model. The teacher model is used as a baseline to see how much the student represents the state-of-the-art accuracy. The independent student model is trained directly on targets without applying
any form of knowledge transfer, and is used as a baseline to see how much improvement the knowledge transfer method brings to the dependent student.

% We evaluate the effectiveness of our proposed knowledge transfer on two important aspects. First, the student model can learn incrementally from locally available data without suffering from catastrophic forgetting using the distilled knowledge from the teacher model. Second, the student model can rely on the distilled knowledge from the teacher to train and then classify unseen categories of the student model.

\begin{table}[t]
	\centering
	\caption{Convergence time of compressed ResNet models on CIFAR-10, and VGG-16 models on Caltech 101.}
	\label{tab:convergence_time_of_resnet}
	\begin{tabular}{p{1.45cm}p{0.85cm}p{0.8cm}p{1.14cm}p{0.85cm}p{1.2cm}}
		\toprule
		Model Name & Indepen.  & Depend. (Our)& Depend. (FitNet)&Speedup (Our)&Speedup (FitNet)\\
		\hline
		ResNet 1 & 70.20K& 12.09K& 69.42K& \textbf{5.81$\times$}& 1.01$\times$ \\
        ResNet 2 & 69.42K& 12.87K& 69.03K& 5.39$\times$& 1.01$\times$ \\
        ResNet 3 & 69.42K& 15.99K& 69.03K& 4.34$\times$& 1.01$\times$ \\
		ResNet 4 & 70.98K& \textbf{8.97K}&  70.20K& 7.91$\times$& 1.01$\times$ \\
		\hline
		VGG-16 1 & 8.66K& \textbf{1.87K}& 6.08K& \textbf{4.63}$\times$& 1.42$\times$ \\
        VGG-16 2 & 3.98K& 3.28K         & 3.28K& 1.21$\times$& 1.21$\times$ \\
        VGG-16 3 & 3.28K& \textbf{1.87K}& 3.28K& 1.75$\times$& 1.00$\times$ \\
        VGG-16 4 & 3.28K& \textbf{1.87K}& 4.45K& 1.75$\times$& 0.74$\times$ \\
		\bottomrule
	\end{tabular}
\vspace{-0.2in}
\end{table}

\subsubsection{Incremental learning}
We first evaluate on the incremental learning tasks using both CIFAR-10 and ImageNet dataset with ResNet (WRN-28-10 and ResNet-34) as the teacher models, respectively. Our goal is to allow the student model, which is compressed from the teacher model, to learn one or multiple new, locally available categories without forgetting those old categories that the teacher is trained with. Re-training the whole model on edge devices is infeasible due to their limited computing resources; fine-tuning the on-device model can significantly reduce the training time, but its performance on old categories degrade severely. Our proposed method allows the teacher model to provide distilled knowledge to guide the student model and prevent it from forgetting about the old categories. In the experiment, we first pre-trained with 9 categories of data and try to learn a new category.

Table~\ref{tab:single_task_accuracy_cifar} lists the accuracy for incremental learning using CIFAR-10. As expected, the independent student (w/o KT) cannot classify any of the old categories any more even though it performs well on the new category. In contract, our dependent students (w/ KT) not only perform well on the new category ($>$90\% accuracy) but also retains a good level of accuracy for classifying the old categories. The distilled knowledge from the teacher model significantly alleviates the catastrophic forgetting, when the student model learns the new categories incrementally. Our proposed method also works well for highly compressed models. ResNet 4 (w/ KT), with a compression ratio of 160X, achieves an accuracy only 2\% lower than that of the ResNet 1 model. 

% The oracle model (ResNet 1 Oracle) achieves an average accuracy of 94.35\%
We then compare our method with the related work~\cite{li2019rilod} by applying its knowledge transfer method to two models, ResNet d28w10, the original, uncompressed model used by RILOD, which has a compression ratio of 18X. 
On d28w10, RILOD's accuracy is 7.66\% lower than our ResNet 2 (w/ KT) even though it has \bl{9.58M} more parameters, indicating that our proposed method can better solve the incremental learning problem. On ResNet 1, our accuracy improvement is even more significant (20.84\%). The results demonstrate that our proposed knowledge transfer method can better support incremental training on edge devices, especially for small models suited for edge deployment. 

Table~\ref{tab:single_task_accuracy_imagenet} lists the results for incremental learning on ImageNet. We sampled 10 random categories from the ImageNet dataset as our experiment dataset. We can observe similar results as the aforementioned CIFAR-10 results. The independent student (w/o KT) achieves only 11.1\% accuracy on the 9 old categories after learning one new category. Our dependent student (ResNet-5 (w/ KT)) achieves 57.33\% accuracy on the 9 old categories and 72\% on the new category. Even the highly compressed model, ResNet 8, is only 2\% less accurate than the ResNet 5 model.
% The oracle model that is trained with all 10 categories of data achieves X\% accuracy whereas

\begin{table}[t]
	\caption{Top-1 accuracy on CIFAR-10 for single-task incremental learning using 9+1 categories of data.}
	\label{tab:single_task_accuracy_cifar}
	\begin{tabular}{llll}
		\toprule
		Model    & 9 old & 1 new & Avg Acc  \\
		\hline
% 		ResNet 1 Oracle     &        92.99 &    95.7  &    94.35                    \\
		ResNet 1 w/o KT     &        0             &           \textbf{100}            &    10   \\
		ResNet 1 (w/ KT) & 62.83                     & 93.5                     & 78.16                      \\
		
		ResNet 2 (w/ KT) & 64.12                     & 92.3                     & \textbf{78.21}                      \\
		ResNet 3 (w/ KT) & 61.5                      & 93.1                    & 77.3                       \\
		ResNet 4 (w/ KT) & 57.02                     & 95                      & 76.01                      \\
		ResNet 1 (RILOD)    & 38.34                &  76.3                    & 57.32                                                     \\
		d28w10 (RILOD) & \textbf{83.41}            & 57.6                     & 70.5      \\
		\bottomrule
	\end{tabular}
	\vspace{-0.1in}
\end{table}

\begin{table}[t]
	\caption{Top-1 accuracy on ImageNet for single-task incremental learning using 9+1 categories of data.}
	\label{tab:single_task_accuracy_imagenet}
	\begin{tabular}{llll}
		\toprule
		Model    & 9 old & 1 new & Avg Accuracy  \\
		\hline
% 		Oracle        & X                          &    X           &  X                                  \\
		ResNet 5 w/o KT        & 11.1                          &    \textbf{100}           &    19.9                                    \\
		ResNet 5 (w/ KT) & 57.33                     & 72                    & 64                              \\
		ResNet 5 (RILOD)  & 51.78    & 62  & 56.89              \\   
		ResNet-18 (RILOD)  & 33    & 98  & 65.5              \\   

		ResNet 6 (w/ KT) & \textbf{62}                     & 68                    & 65                              \\
		ResNet 7 (w/ KT) & 57.7                     & 78                    & \textbf{67.89}                               \\
		ResNet 8 (w/ KT) & 56.4                     & 76                    & 66.2                              \\
		\bottomrule
	\end{tabular}
	\vspace{-0.2in}
\end{table}

\begin{table}[t]
	\centering
	\caption{Runtime on ImageNet for single-task incremental learning using 9+1 categories of data on the edge. 
% 	Training time is measured by the total runtime required to converge. Inference time is the time needed for classifying one image.
	}
	\label{tab:single_task_runtime_cifar}
	\begin{tabular}{p{2.9cm}p{1.2cm}p{1.3cm}p{1.2cm}}
		\toprule
		Model Name              & Trainable Para.      & Training Time (s) & Inference Time (s)\\
		\hline
		ResNet-18 (RILOD)   & 11.69   & 1200           & 8.7 \\
		ResNet 5                & 3.51    & 360            & 7.2 \\
		ResNet 6                & 3.07    & 315            & 5.3    \\
		ResNet 7                & 2.46    & 253            & 3.7 \\
		ResNet 8                & 1.76    & \textbf{181}   & \textbf{2.5} \\
		\bottomrule
	\end{tabular}
    \vspace{-0.2in}
\end{table}

In addition to improving accuracy, by using compressed models for incremental training on edge devices, our approach also runs faster than the original RILDO approach which uses a large, uncompressed model (ResNet-18) on devices. 
Table~\ref{tab:single_task_runtime_cifar} compares the training time and inference time of our compressed ResNet models between these two approaches. Training time is measured by the total runtime required to converge, and inference time is measured by the time needed for classifying one batch of images.

All our compressed models can converge within three to six minutes on the edge whereas RILOD needs 20 minutes. In addition, the inference time of our models (from 2.5s to 7.2s) is also shorter than that of RILOD (8.7s). The above results demonstrate the importance of our model compression and knowledge transfer techniques in improving both accuracy and runtime performance of incremental learning on edge devices.

\subsubsection{Classifying Unseen Categories}
We then study the performance of the compressed student models when presented with data from classes that they are never trained with. In this experiment, we train our compressed ResNet models only with 10,000 images from two categories of the original training dataset (CIFAR-10), which has ten categories and 5,000 images each. Then we test them with 8,000 images from the other eight categories that they are never trained with. 

Table~\ref{tab:KT accuracy for unseen classes} shows the accuracy of the independent student and dependent student models of classifying the unseen eight categories of images. For all of the compressed ResNet models, the accuracy of the independent students is all zero, showing that they cannot classify the unseen categories. Since they never saw those categories during training, no features were learned. However, our proposed knowledge transfer method allows the student models to achieve an accuracy of at least 60\%. ResNet 1 and ResNet 2 achieve an accuracy of 78.92\% and 75.72\% respectively, even though they are trained with only such a small dataset including only two categories of images. FitNet cannot help in this case and its accuracy is zero on all of its dependent students. The reason is likely that FitNet learns the features generated from the teacher, instead of the process of solving a problem, so its knowledge transfer works only when the target tasks of the teacher and student models are similar.

\begin{table}[t]
	\centering
% 	\caption{Top-1 accuracy of knowledge transfer methods for classifying unseen categories on CIFAR-10.}
	\caption{Top-1 accuracy of knowledge transfer methods for classifying unseen categories on CIFAR-10. 
% 	Training dataset includes 2 categories (cat and dog), each of which has 5,000 images. Testing dataset includes the other 8 categories, each of which has 1,000 images. Accuracy of the teacher model on the eight categories is 97.46\%.
	}
	\label{tab:KT accuracy for unseen classes}
	\begin{tabular}{p{2.0cm}p{2.0cm}p{1.5cm}p{1.5cm}}
		\toprule
		Model Name & Independent (Baseline) & Dependent (Our KT)& Dependent (FitNet)\\
		\hline
		ResNet 1 & 0.00\% & \textbf{78.92\%} & 0.00\% \\
		ResNet 2 & 0.00\% & 75.72\% & 0.00\% \\
		ResNet 3 & 0.00\% & 70.40\% & 0.00\% \\
		ResNet 4 & 0.00\% & 61.96\% & 0.00\% \\
		\bottomrule
	\end{tabular}
	\vspace{-0.2in}
\end{table}
% !TEX root = main.tex
\section{Related Works}\label{related}
Model compression techniques can be broadly classified into three categories, weight sharing, quantization, and pruning techniques. Weight sharing reduces the occupied memory by using the same set of weights to represent more than one transformations~\cite{han2015deep, chen2015compressing}. Quantization reduces the size of the model by shrinking the number of bits needed for storing the weights~\cite{han2015deep,kadetotad2016efficient}. Pruning removes redundant weights or neurons while minimizing accuracy loss. Han proposed to remove weights below a particular threshold~\cite{han2015deep}. Li proposed to prune filters with the lowest absolute weight sum~\cite{li2016pruning}. But the above related works \bl{focused only} on creating compressed models for efficient inference, and did not consider how the compression methods affect the training process, and \bl{how to address the accuracy loss caused by compression}. More recently, Polino et al. proposed Quantized Distillation, which \bl{leverages} quantization and distillation jointly during the training process of the smaller model~\cite{polino2018model}. %As shown in Section~\ref{model_compression_results}, our filter pruning based model compression method alone without knowledge transfer enables the compressed model achieves a comparable Top-1 accuracy as Quantized Distillation (93.68\% vs 94.73\%), but with much less parameters (1.42M vs 9.66M). Moreover, our knowledge transfer method allows the compressed model to learn from new data available on the edge, which is not possible with the related work.     

% More recently, Polino et al. proposed Quantized Distillation, which \bl{leverages} quantization and distillation jointly during the training process of the smaller model~\cite{polino2018model}. As shown in Section~\ref{model_compression_results}, our filter pruning based model compression method alone without knowledge transfer enables the compressed model achieves a comparable Top-1 accuracy as Quantized Distillation (93.68\% vs 94.73\%), but with much less parameters (1.42M vs 9.66M). Moreover, our knowledge transfer method allows the compressed model to learn from new data available on the edge, which is not possible with the related work.     
 
Existing knowledge transfer techniques can be broadly classified into three categories, including transferring hard logits, transferring soft logits, and transferring intermediate representations. Ba et al. proposed hard-logits-based knowledge transfer technique~\cite{ba2014deep}, which minimizes squared difference (RMSE) between the logits of the teacher and the shallow student. Hinton et al. introduced transferring soft logits~\cite{hinton2015distilling} where the student minimizes the sum of two objective functions: (1) cross entropy loss between the soft logits, and (2) cross entropy loss between the softmax output and correct labels of the dataset. Romero et al. proposed FitNet, which extended transferring soft logits by using not only the soft outputs but also the intermediate representations learned by the teacher~\cite{venkatesan2016diving}. More recently, FSP matrix transfer~\cite{yim2017gift}, attention transfer~\cite{zagoruyko2016paying}, and factor transfer~\cite{kim2018paraphrasing} were proposed, which are also based on transferring the intermediate representations.

However, these knowledge transfer methods have the following limitations: 1) they still require large student models that are not fit for resource-constrained devices. For example, the student model used in RILOD~\cite{li2019rilod} has the same architecture with the teacher model; FitNet is thinner but not shallower, and FSP is shallower but not thinner. In comparison, we integrate proposed filter based model compression method with knowledge transfer method, enabling the student model both shallower and thinner than the teacher, which is important for deployment on resource-constrained devices; 2) they enable their student models classifying only the categories that the models are trained with, whereas our method also allows the students to classify unseen categories with a good accuracy; \bl{3) they did not update the student incrementally in real time, whereas our method enables the students learning new categories within 181 seconds on the edge while remaining a good accuracy on the old categories.}
% !TEX root = main.tex
\section{Conclusions}\label{conclusions}
This paper provides a novel solution to deploying and training state-of-the-art models on resourced-constrained edge devices. Our results show that, by pruning similar layers in a model and the filters that produce weak activation patterns in each layer, complex DNNs can be reduced to both shallower and thinner networks, suitable for deployment on devices but without much loss of accuracy. In addition to reducing the size, our solution allows our compressed models to converge the training within three to six minutes on the edge. Our results also show that such compressed models can also learn incrementally on new data without forgetting the old categories. In addition, our results show that transferring the problem solving process is much more effective than letting the student simply mimic teacher's intermediate results. It allows the on-device model to be trained with both good accuracy and speed, without relying on the input's true labels, and to recognize unseen categories.

% % In the unusual situation where you want a paper to appear in the
% % references without citing it in the main text, use \nocite
\nocite{langley00}

\bibliography{reference.bib}
\bibliographystyle{icml2020}
\end{document}